# ATTENTION-BASED SEQUENCE-TO-SEQUENCE MODEL FOR SPEECH RECOGNITION: DEVELOPMENT OF STATE-OF-THE-ART SYSTEM ON LIBRISPEECH AND ITS APPLICATION TO NON-NATIVE ENGLISH

*Yan Yin, Ramon Prieto, Bin Wang, Jianwei Zhou, Yiwei Gu, Yang Liu, Hui Lin*

LAIX Inc.
{yin.yan, Ramon.Prieto, engine.wang, yang.liu}@liulishuo.com

## ABSTRACT

Recent research has shown that attention-based sequence-to-sequence models such as Listen, Attend, and Spell (LAS) yield comparable results to state-of-the-art ASR systems on various tasks. In this paper, we describe the development of such a system and demonstrate its performance on two tasks: first we achieve a new state-of-the-art word error rate of 3.43% on the test clean subset of LibriSpeech English data; second on non-native English speech, including both read speech and spontaneous speech, we obtain very competitive results compared to a conventional system built with the most updated Kaldi recipe.

*Index Terms—* attention model, sequence-to-sequence model, end to end speech recognition, non-native English

## 1. INTRODUCTION

Conventional neural network based speech recognition systems [1-4] using a hybrid HMM/NN approach require a lexicon, train acoustic models and language models separately, and use sophisticated beam search decoding with complicated search space optimization. Recently there has been growing interest in building sequence-to-sequence speech recognition systems that use a consolidated neural network framework to subsume all the necessary components (pronunciation, acoustic and language models) of conventional speech recognition systems. A variety of sequence-to-sequence models have been explored [5-8], such as RNN transducer [5], Listen, Attend, and Spell (LAS) [6]. Training such models is less complicated than conventional ASR systems. Since such model can directly operate on character or sub-words, pronunciation lexicon is not needed, nor decision trees or alignment. Recent studies [9-11] have shown that such sequence-to-sequence ASR systems can achieve very competitive results compared to conventional state-of-the-art systems on a variety of tasks from publicly available corpus (LibriSpeech, Switchboard) to real industry data such as Google voice search.

This work aims to evaluate the effectiveness of sequence-to-sequence ASR systems for non-native speech recognition. A motivating reason for applying such methods to non-native speech is because of the large pronunciation variation of non-native speakers or language learners. We expect it is more effective to jointly learn the lexicon with acoustic models. In this work, we develop a sequence-to-sequence system based on LAS, following previous work [12]. First we demonstrate its performance on Librispeech data, with a new state-of-the-art result (WER 3.43% on clean Test). Then we apply this system to our non-native English data, showing comparable WER results to a Kaldi-based system.

## 2. SYSTEM DESCRIPTION

### 2.1 Vanilla LAS Model

The basic LAS model consists of two sub-modules: the listener and the speller. The listener is an acoustic model encoder, which takes the input speech features $x$, and transforms it into high level representation $h = (h_1, \dots, h_U)$ with $U \ll T$ ($T$ is the original frame number). For speech data, the length of the input feature $x$ can be hundreds to thousands of frames long, and applying a BLSTM on such a long sequence converges slowly and yields poor results. This can be circumvented by using pyramid BLSTM (pBLSTM). In each successive stacked pBLSTM layer, the time resolution is reduced by a factor of 2. This is done by concatenating the outputs of two consecutive steps of each layer before feeding it to the next layer. The speller is an attention-based decoder, which consumes $h$ and produces a probability distribution of the target sequence $p(y|x)$. The attention mechanism in speller determines, at each time step $i$, which encoder output features in $h$ should be attended to, and then generates a context vector $c_i$. The speller then takes the context vector $c_i$ along with the previous prediction $y_{i-1}$, and generates a probability distribution $p(y_i|x, y < i)$.

### 2.2 Improvements over LAS

#### 2.2.1 Location-Aware Attention

When computing the context vector, the alignment or the attention weight $\alpha_{ij}$ is obtained by:

$$e_{ij} = w^T \tanh(W s_{i-1} + V h_j + b) \quad (1)$$
$$\alpha_{ij} = \exp(e_{ij}) / \sum_{j=1}^{U} \exp(e_{ij}) \quad (2)$$

The main limitation of such a scheme is that identical or very similar elements of h are scored equally regardless of their positions in the sequence. Alternatively, the location-based attention mechanism proposed in [13] takes into consideration the alignment history when computing the current time step's alignment. Such alignment history information tells the attention mechanism which parts of the speech have been attended, thus allowing attention to focus on the right region at the current time step. More specifically, convolution is applied to alignment history:

$$f_i = F * \alpha_{k<i} \quad (3)$$
$$e_{ij} = w^T \tanh(Ws_{i-1} + Vh_j + Uf_{ij} + b) \quad (4)$$

Either the previous time step's alignment or the accumulated alignments can be used as alignment history. In our experiments, we did not observe much performance difference between them.

### 2.2.2 Scheduled Sampling

The so-called teacher forcing feeds the ground truth as the previous label prediction and uses it in training, which helps the speller to learn quickly at the beginning. However, this introduces the mismatch between training and inference. Scheduled sampling, as used in [9] [14], samples from the probability distributions of the previous prediction and then feeds the sampled results for next label prediction. This process helps reduce the gap between training and inference. Two scheduled sampling strategies have been investigated in the past. In [9], a small sampling probability is used at the early stage of training, and then linearly ramps up as training proceeds. In [10], scheduled sampling is adopted once the cross-entropy (CE) loss on the validation set is saturated. We compared the two strategies and observed slightly better performance from the latter one in our experiments.

### 2.2.3 Label Smoothing

In cross-entropy training, the ground truth is fed as a one-hot vector, and by minimizing cross-entropy loss, the logit corresponding to the ground truth label becomes much larger than all the other logits. This may result in overfitting to the training data. Similar to [9] [15], we use label smoothing, which smooths the ground-truth label distribution with a uniform distribution over all the labels other than the reference one. This makes the model generalize better and prevents it from being too confident about its predictions.

### 2.2.4 Stabilized Training

Several strategies are adopted to stabilize the training. First, the new-bob learning rate policy is used. During training, we decay the learning rate when the validation loss stops dropping. Second, similar to [9], we use learning rate warmup and gradient norm tracker. For learning rate warmup, we start training with a very small learning rate and gradually increase it as training proceeds. This is done in the early training epochs. For gradient norm tracker, we keep track of the gradient norm distribution, and clip the gradients when their norm falls at the tail of the norm distribution. We evaluated norm tracking at global and per-variable level and did not observe much difference. Moreover, we noticed that even with gradient norm tracking, sometimes we still see large gradient norms, which may cause training instability. We further adopt a static norm threshold and combine it with gradient norm tracker to improve the training stability. These are crucial especially for multi-GPU training.

### 2.2.5 Word Piece Model

Recent work has shown that using longer units such as word pieces (WP) performs significantly better than characters or graphemes [9,11]. In this work we also adopt WP models. We use byte-pair encoding (BPE) [16] to create WPs from the text corpus. Beam search decoding outputs WP level hypotheses, which are then merged into words.

### 2.2.6 Discriminative Training

In this work, we adopt minimum expected word error rate (MWER) training [9] using n-best lists. The objective is to minimize the expected number of word errors $W(y, y^*)$. To stabilize training, the MWER objective function is interpolated with standard cross-entropy (CE) loss function,

$$L_{mwer} = E_{p(y|x)} W(y, y^*) + \lambda L_{ce} \quad (5)$$

We approximate the loss function by restricting the summation to an n-best list of hypotheses, and compute the weighted sum of the normalized word error (by ground truth length $L$) for the n-best list. The LAS output probability scores renormalized over the n-best list, $P^*(y_i|x)$, are used as weights. Since the score of the top-1 hypothesis is usually dominant, we apply exponential scaling to the scores of the n-best hypotheses with a tunable scaling factor less than one, and then re-normalize the scores such that the sum of the n-best scores equals to one.

$$L_{mwer}^{nbest} = \sum_{y_i \in NBest(x,N)} \frac{1}{L} P^*(y_i|x) W(y_i, y^*) + \lambda L_{ce} \quad (6)$$

We start MWER training after the CE training converges. This is similar to conventional ASR systems, where discriminative training (e.g., state-level minimum Bayes risk (sMBR) [17, 18]) is usually conducted on top of CE training.

### 2.3 Data Augmentation

In our work, we perform speed perturbation-based data augmentation. To modify the speed of a signal we just resample the signal. The *speed* function of *Sox* is used for this. Two additional copies of the original training data are created by modifying the speed to 90% and 110% of the original rate. Since including all the augmented data in training increases training time, we only use them in the discriminative training step after the initial CE training is done.

### 2.4 External Language Model (LM) Combination

While the LAS speller learns a build-in LM, it is only exposed to the training transcripts. An external LM, on the other hand,

can leverage large amounts of additional text data, thus may still improve the LAS performance when the training size is not large. In our work, we investigate using external LMs in both first pass and second pass decoding.

**(A) First pass shallow fusion**

In shallow fusion [19], the external LSTM LM is incorporated via log-linear interpolation at inference time in beam search:

$$y^* = argmax_y logp(y|x) + \lambda logp_{lm}(y) \qquad (7)$$

In our work, we implemented shallow fusion with word piece LSTM LM. Specifically, at the beginning of each utterance we initialize the state of LSTM LM to zero and feed the end-of-sentence symbol to the LSTM LM input. At each step in the beam search decoder and for each word piece, we add the LSTM LM log-probability to the LAS log-probability of that word piece. For shallow fusion we found that compensating for the length bias of the models with a length penalty weight as defined in [22] achieved the best results. We applied this length penalty weight at each step during beam search decoding.

**(B) Second pass rescoring**

To use external LM for second pass rescoring, we do beam search decoding to generate n-best hypotheses and their corresponding probability scores from the LAS model. Then we rescore the n-best hypotheses by interpolating the LAS probability scores and external LM scores. In our work, both n-gram, word and word-piece LSTM LMs are investigated for rescoring.

## 3. EXPERIMENTS

We conduct experiments on two tasks. First we show our system is very competitive, achieving the best results on LibriSpeech. Then we present results on our own non-native English data sets, spoken by Chinese.

### 3.1 LibriSpeech Results

LibriSpeech training data consists of about 1000 hours of read audio books [21]. The dev and test sets are split into clean and other subsets. In our experiments, we only used clean subsets for both Dev and Test. Our system is tuned based on the WER on the Dev set, and the final optimized system is evaluated on Test.

We use 40-dimention static filter-bank features. The encoder network architecture consists of 3 pyramid BLSTM layers with 1024 hidden units in each direction. The decoder is a 2-layer LSTM with 512 hidden units per layer. Networks are trained with the cross-entropy criterion first, and then the model is used as the seed for discriminative training. We use 4-gpu synchronized training. TensorFlow [20] is used for all the experiments.

Table 1 shows the contributions of different components in the LAS model. No external LM is used in these experiments.

Using a character model as the baseline, we can see that stabilized training provides a significant 13% relative WER reduction (WERR). The new-bob learning rate policy is based on an initial learning rate of 0.002 and decay rate of 0.9; learning rate warmup starts from a very small value of 0.0002 and ramps up to 0.002 in 2 epochs. For gradient norm tracking, a moving average with a decay rate 0.95 is used for mean computation. A standard deviation factor 2.0 is used to define the long-tail region. Gradients with norms falling into the long-tail region are clipped with norm mean as maximum norm. Static gradient norm 5.0 is combined with gradient norm tracker to further control the stability.

|  | Dev | Test | WERR |
| --- | --- | --- | --- |
| Char baseline | 6.57 | 6.61 | - |
| + ST | 5.72 | 5.75 | 13.0% |
| WP + ST | 5.29 | 5.32 | 7.5% |
| + LA-attention | 4.70 | 4.83 | 9.2% |
| + LS | 4.37 | 4.62 | 4.3% |
| + SS | 4.31 | 4.51 | 2.2% |
| + DT | 4.29 | 4.40 | 2.4% |
| + DA | 4.21 | 4.34 | 1.4% |

Table 1: WER results (%) on Librispeech Dev and Test sets without using external LMs. ST: stabilized training, WP: word piece, LA: location-aware attention, LS: label smoothing, SS: scheduled sampling, DT: discriminative training, DA: data augmentation.

Our WP model is based on about 500 word pieces, which gives 7.5% relative WERR over the character model. For location-aware attention (which yields additional 9.2% WERR), we do convolution with 20 filters and window size 100. Our label smoothing experiments with a smoothing factor 0.01 achieves the best results with 4.3% WERR. For scheduled sampling, we increase the sampling probability to 0.2 after the model trained with sampling probability 0.1 is saturated, which yields another 2.2% WERR. Using our best CE trained model as the seed, we further did MWER discriminative training (DT) in combination with CE loss (its weight is 0.01) based on N-best list (N=4) and obtain 2.4% WERR. Finally, running DT with data augmentation gives an extra 1.4% WERR. Again, note that data augmentation is only used in DT to reduce the training cost. We expect to see more gain using augmented data from scratch for model training. Our best model without using external LMs achieves 4.34% WER on the test clean subset.

Table 2 shows the results using different ways to incorporate external LMs. For N-gram LM rescoring, we used the full 4-gram LM released in [21], which were trained using modified Kneser-Ney smoothing and no pruning. In addition, we heavily penalized the UNK token unigram probability and eliminated N-grams of order greater than one that contained this UNK token. Our WP LSTM LM consists of 2 layers and 1024 nodes per layer. No extra projection layers were used since the number of word pieces is very small. We did not use learning rate decay or dropout. Training was done with 8 GPUs, with gradient averaging across the GPUs and clipping gradients by a global norm threshold. A word LSTM LM with 256 embedding size, 2 hidden layers and 1024 hidden units per layer are also trained on the extra LM training text. After mapping the words that occurred few than 4 times to the UNK token, we get a vocabulary including about 348K words. Then adaptive softmax [24] is used to reduce the time and memory cost. This model is trained with Adam [25] and a fixed learning rate 0.001.

We used the dev set to tune, not only the LM weight, but also the length penalty weight in the beam search decoder. Rescoring is done on N-best list (N=16) generated from LAS beam search decoder. As can be seen from Table 2, incorporating the external LM significantly reduces WER. Our best system with external LSTM WP LM shallow fusion achieves WER of 3.17% on dev clean set, and 3.43% on test clean set. This is by far the best published result on this data set. It is worth pointing out that we just noticed a new result on Librispeech data [26] while this paper is being submitted. That system uses a more complicated network and data augmentation.

|  | Dev | Test | WERR |
|---|---|---|---|
| No LM | 4.21 | 4.34 | - |
| 4-Gram LM rescore | 3.52 | 3.76 | 13.4% |
| LSTM word LM rescore | 3.48 | 3.68 | 15.2% |
| LSTM WP LM rescore | 3.33 | 3.57 | 17.7% |
| LSTM WP LM SF | **3.17** | **3.43** | 21.0% |

Table 2: WER results (%) on Librispeech Dev and Test sets with external LM incorporated via shallow fusion (SF) and second pass rescoring.

### 3.2 Non-native English Results

In this experiment we evaluate how the sequence-to-sequence model performs on our in-house dataset of non-native English data, spoken by language learners whose native language is Chinese. This data is mostly collected from Liulishuo language learning App. Two test sets are used in this study:

- **Read speech** (4.3 hours): this is user reading a given sentence prompt (after listening to a native speaker's speech).
- **Spontaneous speech** (6.4 hours): this is the IELTS speaking practice test speech.

The training data consists of over 3K hours of speech from the following four different datasets: non-native read speech, same style as that in the test set (about 2700 hours), spontaneous speech from IELTS test practices (about 300 hours), read speech from native speakers (about 120 hours), and a public speech recognition training dataset from TED talks (about 114 hours) [23].

We did not optimize the system for different domains and used the same one for the two test conditions. Since the spontaneous speech is a small portion in training, we perform data augmentation to that dataset in order to balance the speech from different domains. This results in about 4K hours speech for model training. We randomly selected 5% speech from each set to generate a dev set and used the rest as the training set. DT/MWER is also run using only the spontaneous speech portion. All the hyper-parameters are tuned on the dev set. The configuration of this LAS system is quite similar to that used for Librispeech, with the following few differences: the learning rate in new-bob learning was initialized to 0.001 and decayed with a rate of 0.5; the sampling probability of scheduled sampling was fixed to 0.1 during training; and shallow fusion was not used in this experiment.

The WERs on each test set are shown in Table 3. The baseline is a Kaldi-based system, where the acoustic model was trained using the public TDNN-F recipes. A trigram word LM is used in the first pass decoding to generate the n-best list and then a word LSTM LM is used to rescore the n-best. All the LMs are trained on an external training corpus including about 1 billion words. 16-best list generated by LAS model is also rescored using these LMs. We can see that the Kaldi system benefited more from LM rescoring, and that there is limited gain from LM rescoring for LAS in this experiment. This may be because that the decoder in LAS is a better match for the test sets than LMs that are trained on the large external corpus, where most text is in news domain. LAS without using any external LM can even outperform the best Kaldi system (with the external LSTM LM) for spontaneous test set and Kaldi with external trigram LM for read speech. Overall, we conclude that our LAS system has comparable performance to the Kaldi system: slightly worse than Kaldi on read speech, and better on the spontaneous test set. These initial results demonstrate the potential of such sequence-to-sequence models in non-native English ASR tasks.

|  | Read | Spontaneous |
|---|---|---|
| Kaldi baseline | - | - |
| + trigram | 11.47 | 16.47 |
| + LSTM | 9.95 | 15.44 |
| LAS | 10.44 | 15.16 |
| + trigram | 10.43 | 14.80 |
| + LSTM | 10.41 | 14.77 |

Table 3: WERs (%) on Liulishuo non-native English tasks.

### 4. CONCLUSIONS

We have developed an attention-based sequence to sequence ASR system, and achieved a WER of 3.43% on test clean set of LibriSpeech task. Furthermore, we evaluated our model on our non-native English task, and showed it achieved very competitive results compared to systems trained based on the most recent Kaldi recipe. In our future work, we will continue our efforts on applying this sequence-to-sequence model to the non-native English task and address some special issues from this task.

## 5. ACKNOWLEDGEMENT

The authors thank Patrick Nguyen from Google and Jia Cui from Tencent for helpful discussions.